\pdfoutput=1

\documentclass[11pt]{article}

\usepackage[preprint]{acl}

\usepackage{times}
\usepackage{latexsym}

\usepackage[T1]{fontenc}
\usepackage{xspace}
\usepackage{adjustbox}
\usepackage{amsmath,amssymb,amsthm,bbm}
\usepackage{xcolor}
\usepackage{xspace}

\DeclareFontFamily{U}{mathx}{\hyphenchar\font45}
\DeclareFontShape{U}{mathx}{m}{n}{<-> mathx10}{}
\DeclareSymbolFont{mathx}{U}{mathx}{m}{n}
\DeclareMathAccent{\widebar}{0}{mathx}{"73}

\newcommand{\R}{\ensuremath{\mathbb{R}}}

\usepackage{mathtools}

\definecolor{light-gray}{gray}{0.80}

\definecolor{darkred}{rgb}{0.64, 0.0, 0.0}

\theoremstyle{definition}

\newtheorem*{thm*}{Theorem}

\newenvironment{itemizesquish}{\begin{list}{\labelitemi}{\setlength{\itemsep}{-0.2em}\setlength{\labelwidth}{0.5em}\setlength{\leftmargin}{\labelwidth}
\addtolength{\leftmargin}{\labelsep}}}{\end{list}}

\newcommand{\mingda}[1]{{\color{magenta} [Mingda: {#1}]}}
\newcommand{\todo}[1]{{\color{red} [TODO: {#1}]}}
\newcommand{\wenzheng}[1]{{\color{blue} [Wenzheng: {#1}]}}
\newcommand{\karl}[1]{{\color{brown} [Karl: {#1}]}}
\newcommand{\scott}[1]{{\color{violet} [Scott: {#1}]}}
\newcommand{\vic}[1]{{\color{teal} [Victoria: {#1}]}}

\renewcommand{\mingda}[1]{}
\renewcommand{\todo}[1]{}
\renewcommand{\wenzheng}[1]{}
\renewcommand{\karl}[1]{}
\renewcommand{\scott}[1]{}
\renewcommand{\vic}[1]{}

\newcommand{\nq}{NaturalQuestions\xspace}
\newcommand{\nqshort}{NQ\xspace}
\newcommand{\hotpot}{HotpotQA\xspace}
\newcommand{\hotpotshort}{Hopo\xspace}

\newcommand{\wikiqa}{2WikiMultiHopQA\xspace}
\newcommand{\wikiqashort}{2WQA\xspace}
\newcommand{\simpleqa}{SimpleQA\xspace}
\newcommand{\simpleqashort}{SQA\xspace}
\newcommand{\trex}{T-Rex\xspace}
\newcommand{\zsre}{ZsRE\xspace}

\newcommand{\fever}{FEVER\xspace}
\newcommand{\fevershort}{FEV\xspace}
\newcommand{\aida}{AIDA\xspace}
\newcommand{\thiswork}{ImpRAG\xspace}

\newcommand{\oasst}{oasst1\xspace}
\newcommand{\coqa}{CoQA\xspace}
\newcommand{\drop}{DROP\xspace}
\newcommand{\newsqa}{NewsQA\xspace}
\newcommand{\pubmedqa}{PubMedQA\xspace}
\newcommand{\quail}{Quail\xspace}
\newcommand{\squad}{SQuAD\xspace}
\newcommand{\cnndaily}{CNN DailyMail\xspace}

\newcommand{\rait}{RA-IT\xspace}
\newcommand{\radit}{RA-DIT\xspace}
\newcommand{\raditllama}{RA-DIT-Llama\xspace}

\usepackage[utf8]{inputenc}

\usepackage{microtype}

\usepackage{inconsolata}

\usepackage{graphicx}

\title{Learning to Retrieve and Generate Jointly\\for Generalizable Retrieval-Augmented Generation}
\title{Learning to Retrieve and Generate Jointly\\for Query-Free Retrieval-Augmented Generation}
\title{\includegraphics[height=1em]{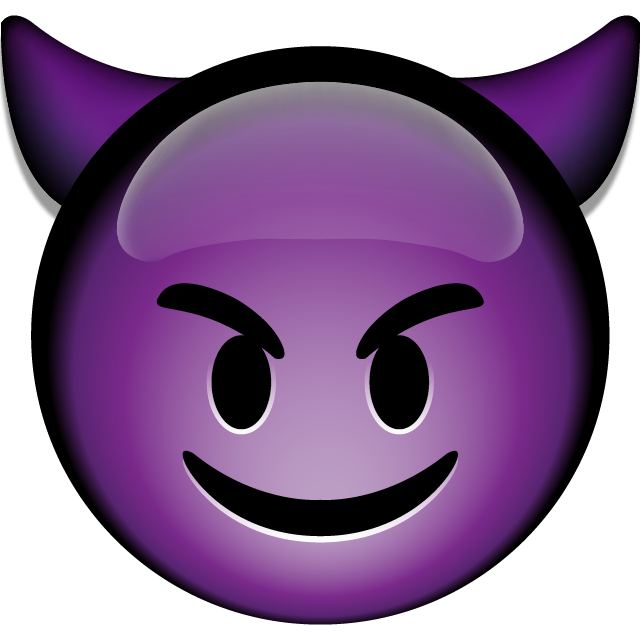} ImpRAG: Retrieval-Augmented Generation with Implicit Queries}

\author{
 \textbf{Wenzheng Zhang\textsuperscript{1}}\thanks{\;\; Work done during an internship at Meta}\hspace{5mm}
 \textbf{Xi Victoria Lin\textsuperscript{2}}\hspace{5mm}
 \textbf{Karl Stratos\textsuperscript{1}}\hspace{5mm}
 \textbf{Wen-tau Yih\textsuperscript{2}}\hspace{5mm}
 \textbf{Mingda Chen\textsuperscript{2}}
\\
\\
 \textsuperscript{1}Rutgers University \hspace{5mm}
 \textsuperscript{2}FAIR, Meta
\\
 \texttt{\{wenzheng.zhang, karl.stratos\}@rutgers.edu} \\
 \texttt{\{victorialin,scottyih,mingdachen\}@meta.com}
}

\begin{document}
\maketitle
\begin{abstract}
Retrieval-Augmented Generation (RAG) systems traditionally treat retrieval and generation as separate processes, requiring explicit textual queries to connect them. This separation can limit the ability of models to generalize across diverse tasks. In this work, we propose a query-free RAG system, named \thiswork, which integrates retrieval and generation into a unified model. \thiswork allows models to implicitly express their information needs, eliminating the need for human-specified queries. By dividing pretrained decoder-only language models into specialized layer groups, \thiswork optimizes retrieval and generation tasks simultaneously. Our approach employs a two-stage inference process, using the same model parameters and forward pass for both retrieval and generation, thereby minimizing the disparity between retrievers and language models. Experiments on 8 knowledge-intensive tasks demonstrate that \thiswork significantly enhances both retrieval and generation performance, with exact match scores increasing by 3.6-11.5 points and retrieval recalls improving by 5.0-23.2 points for unseen tasks with diverse formats, highlighting its effectiveness in enabling models to articulate their own information needs and generalize across tasks. Our analysis underscores the importance of balancing retrieval and generation parameters and leveraging generation perplexities as retrieval training objectives for enhanced performance.


\end{abstract}

\section{Introduction}

\begin{figure}[t]
\begin{center}
    \includegraphics[scale=0.38]{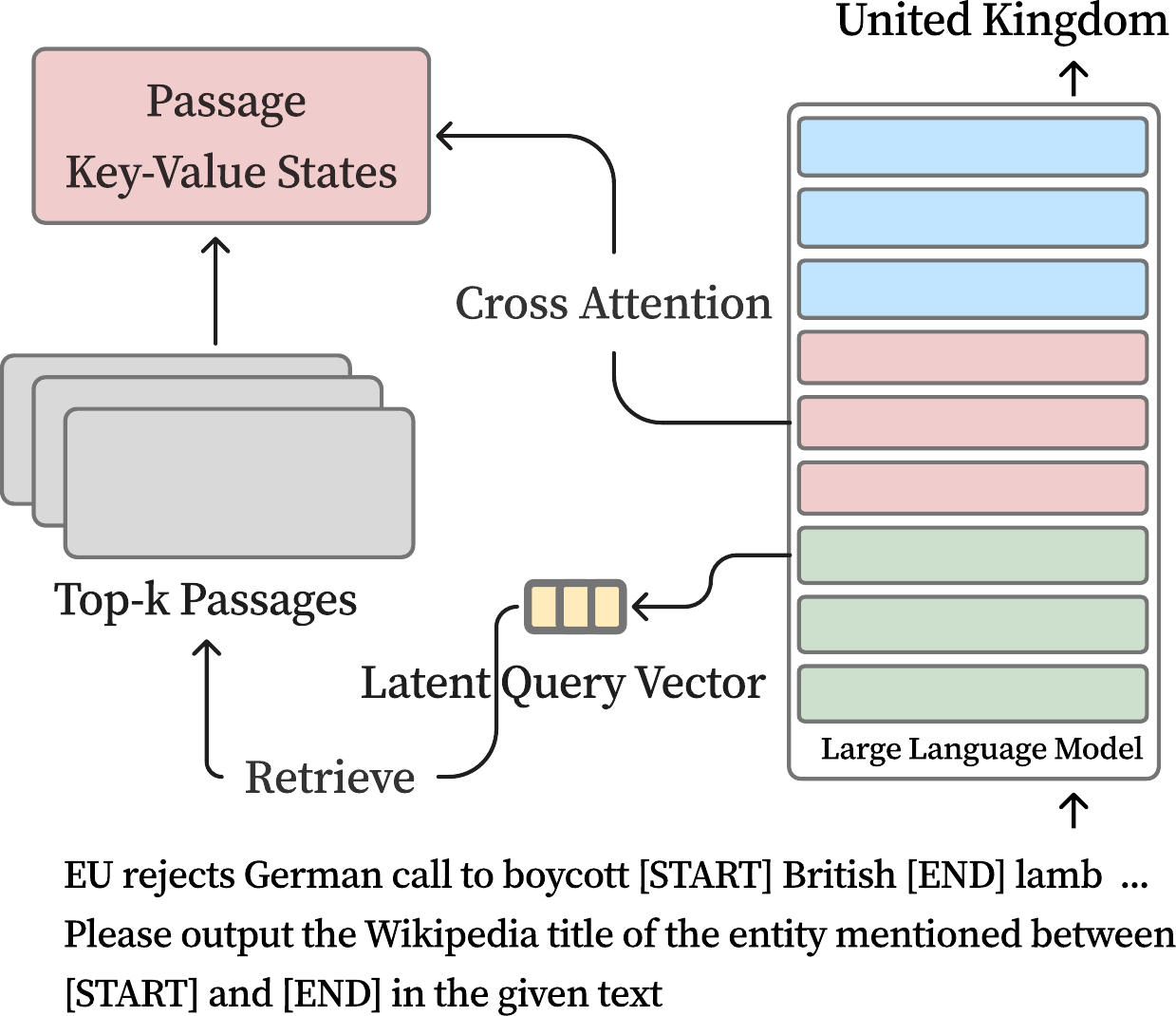}
    \caption{Diagram illustrating the inference process of \thiswork on the entity linking task. We divide decoder-only LLMs into three layer groups for specialized finetuning: bottom (green), middle (red), and top (blue). The bottom layers are optimized for retrieval tasks. The middle and top layers handle the reading of retrieved passages, with cross-attention disabled in the top layers to reduce memory consumption. Standard RAG systems would require a task-specific design of queries (e.g., use the substring ``British'' as the query in the shown example). In contrast, \thiswork uses implicit queries, eliminating the need for explicit specification of queries and allowing models to generalize across unseen tasks with varied formats.
    }\label{fig:arch}
\end{center}
\end{figure}

Retrieval-Augmented Generation (RAG; \citealp{pmlr-v119-guu20a,NEURIPS2020_6b493230,shi-etal-2024-replug}) typically involves two key operations: retrieval and generation. RAG systems retrieve relevant information to enhance generation models, enabling them to respond more effectively to prompts by providing long-tail knowledge or up-to-date information. While effective, traditional approaches often treat retrieval and generation as separate processes, connected by queries.\footnote{In this work, we use the term ``queries'' to refer to textual queries used in an information retrieval setup, unless otherwise specified. This is distinct from queries in the context of self-attention within Transformer architectures.} Consequently, these approaches usually require explicit specification of textual queries. By definition, queries express one's uncertainties; however, in RAG systems, instead of models expressing their information needs, humans must do this for them. This separation can lead to a disconnect between what large language models (LLMs) require and what retrievers assume is necessary. More importantly, it restricts the models' ability to generalize across diverse, unseen tasks during testing. Therefore, in this work, we explore the development of a \textit{query-free} RAG system, enabling models to articulate their own information needs without additional human intervention. 

To achieve this, we introduce \thiswork, a novel approach that integrates retrieval and generation into a unified model and process. This allows models to convey their own information needs implicitly, reducing the need for prior knowledge of test tasks and for humans to formulate explicit textual queries in advance.
At its core, \thiswork aims to enable retrieval capabilities through retrieval heads in self-attention. Building upon pretrained decoder-only language models, \thiswork divides the layers into three groups: the bottom group for retrieval and the middle and top groups for reading and generation.

Figure~\ref{fig:arch} illustrates an example of applying \thiswork to the entity linking task, where models are tasked with linking the mention "British" to an entity in Wikipedia, given the context paragraph. A typical RAG model would require the design of a separate query template, such as using only the mention text, to achieve reasonable retrieval performance. In contrast, \thiswork uses implicit queries and can perform retrieval and generation jointly without the need for additional template design, making it more generalizable.

During training, we optimize two objectives simultaneously: generation loss and retrieval loss. The generation loss is the standard causal language modeling loss, while the retrieval loss first utilizes pseudo labels generated by trained retrievers to warm up the retrieval ability and then self-improves using its own generation log likelihood for the remainder of the training.

At inference time, we employ a two-stage process. First, we embed passages using the bottom layer for retrieval, and then utilize the top layer group to read the retrieved passages and generate the final responses. By leveraging the same forward pass and model parameters for both retrieval and generation, \thiswork reduces the disparity between retrievers and LLMs.

In experiments, we train models on datasets that either require retrieval or do not. The datasets requiring retrieval are used to enhance retrieval performance, while those not requiring retrieval are used to improve models' instruction-following capabilities. We evaluate the models on 8 knowledge-intensive tasks, focusing on different aspects: basic question answering, multihop reasoning, and instruction following. We also establish strong baselines that perform RAG in the retrieve-then-generate paradigm, including \radit~\citep{lin2024radit}, a method that iteratively updates LLMs and retrievers to better align the two.

Our experiments demonstrate that \thiswork achieves slightly better performance on 4 tasks with formats similar to the training tasks, with an improvement of 0.2-0.6 points in exact match scores, all without the need for additional model parameters. Moreover, it significantly outperforms previous approaches on unseen test tasks with more diverse formats, achieving improvements of 3.6-11.5 points in exact match scores and 5.0-23.2 points in retrieval recalls. This highlights the effectiveness of enabling models to articulate their own information needs. Our analysis indicates that carefully selecting layer group boundaries that balance the parameters used for retrieval and generation, using both trained retrievers for warmup and then self-improve by leveraging generation perplexities as retrieval training objectives, and instruction tuning training datasets is crucial for achieving superior performance in \thiswork. Our analysis also reveals that \thiswork is effective in transferring supervision from generation tasks to retrieval tasks, showing the potential of using an unified model architecture for performing retrieval and generation jointly.

\section{Related Work}

There has been a lot of work on using the retrieve-then-generate paradigm for RAG~\citep[\emph{inter alia}]{NEURIPS2020_6b493230,shi-etal-2024-replug}. Many efforts in this line of work have focused on optimizing retrievers using training signals from generation models, and optionally, the reverse \citep{pmlr-v119-guu20a,NEURIPS2020_6b493230,izacard2023atlas,shi-etal-2024-replug}. Although the specifics can differ, these approaches generally utilize distinct models and input templates for the retrieval and generation phases. A closely related study is that of \citet{jiang-etal-2022-retrieval}, which seek to use the same model for retrieval and generation. However, their research primarily focuses on encoder-decoder style models and their models still rely on separate input templates for retrieval and generation. Another related work by \citet{zhang-etal-2024-onegen} explores the use of special tokens for retrieval, but their study emphasizes in-domain task performance rather than unseen task generalization.

This work is also related to research on query formulation in the context of multihop question answering, where previous studies typically generate textual queries by prompting LLMs, followed by retrieval using a separate retriever~\citep[\emph{inter alia}]{lazaridou2022internet,khattab2022demonstrate,press-etal-2023-measuring,trivedi-etal-2023-interleaving,jiang-etal-2023-active}. \citet{chen-etal-2024-shot} enable LLMs to generate textual queries through synthetic data generation. Additionally, this work is connected to memory architectures in RAG \citep{yang2024memory3,lu2024turborag}, which aim to utilize the key-value (KV) caches of LLMs to reduce computational costs, rather than focusing on minimizing the disparities between generation and retrieval. 

Another relevant area of research is instruction tuning for RAG. \citet{lin2024radit} perform instruction tuning for both retrievers and LLMs and then align them through iterative updates. \citet{pmlr-v235-wang24bd} conduct instruction tuning for RETRO-like models \citep{pmlr-v162-borgeaud22a,wang-etal-2023-shall}. \citet{zhang-etal-2024-arl2} align retrievers with LLMs using synthetic data generated by LLMs. Unlike our work, these studies still treat retrieval and generation as separate processes. In a similar vein, researchers have tried to teach retrievers to follow instructions for building general-purpose information retrieval systems~\citep{asai-etal-2023-task,lee-etal-2024-disentangling,oh2024instructir,weller-etal-2025-followir}. Since \thiswork enables its retrieval capabilities by using self-attention, it is related to research on investigating retrieval heads in the context of long context LLMs~\citep{wu2024retrieval}.

\section{Method}
We build on an autoregressive pretrained language model and enable it to perform retrieval and generation jointly.
Our model, \textbf{\thiswork}, is based on the LLaMA 3 family~\citep{grattafiori2024llama}, with architectural modifications to support retrieval and retrieval-augmented generation. 
At a high level, the layer grouping strategy of \thiswork is inspired by the observation that LLMs learn distinct functions at different layers~\citep{zhao-etal-2024-layer}. Consequently, we have designed the layer groups to align with the capabilities required for retrieval-augmented generation, i.e., retrieval and generation. 

\subsection{Architecture}\label{subsec:arch}

\paragraph{Layer Slicing.}
We partition an $N$-layer language model vertically into three groups, as illustrated in Figure~\ref{fig:arch}. The bottom group, spanning layers $0$ to $b$, is denoted as $\mathcal{L_B}$. The middle group, from layer $b$ to $t$, is denoted as $\mathcal{L_M}$, and the top group, from layer $t+1$ to $N{-}1$, as $\mathcal{L_T}$. Note that $\mathcal{L_B}$ and $\mathcal{L_M}$ share layer $b$, while $\mathcal{L_M}$ and $\mathcal{L_T}$ are disjoint. The layer boundaries $b$ and $t$ are treated as hyperparameters and can be tuned to optimize performance across different model configurations.

\paragraph{Bottom Layers as Retriever.}
We repurpose the bottom group $\mathcal{L_B}$ to act as a  \textit{retriever}, in addition to its standard decoder functionality.
Specifically, we apply pooling last-token pooling over the attention query or key states at the final layer $b$ in $\mathcal{L_B}$. Unlike prior work~\citep{muennighoff2024generative}, we retain the original causal attention in the bottom layers rather than enabling bidirectional attention, as we do not observe any performance improvement from this modification.

Let $h_k$ be the number of key attention heads, $g$ the number of query attention groups (as in Grouped-Query Attention~\citep{ainslie-etal-2023-gqa}), and $d_h$ the head dimension. For a query input, we apply last-token pooling by taking the query attention state of its final token, resulting in a grouped query embedding $\mathbf{E}_q^g \in \R^{(h_k g) d_h}$. We then average the attention heads within each group to obtain the final query embedding $\mathbf{E}_q \in \R^{h_k d_h}$.\footnote{Our preliminary results show that taking average heads works slightly better than using individual heads.} Similarly, for each corpus passage, we extract the key attention state of its last token to compute the passage embedding $\mathbf{E}_p \in \R^{h_k d_h}$.
\mingda{Does ``    --query\_pooling\_method "lasttoken\_left\_pad"
    --key\_pooling\_method "lasttoken" '' correspond to last-token pooling instead of mean pooling? If so, can you fix the descriptions here?} \wenzheng{I tried to make the description general, so I didn't specify which pooling we use here. Do you need me to specify that we used last\_token pooling here?}\wenzheng{I just fixed it.}
\mingda{yes, please. It's easier to understand}\mingda{thanks! let's mention that we have some preliminary results to support our choices? By the way, do you still have relevant results to put in the appendix?}\wenzheng{Sorry, I didn't find relevant results about the pooling choices in the spreadsheet. }\mingda{no worries, I think I did the comparison and can put them to the appendix. Do you have results for retrieval heads?}\wenzheng{I have some results for retrieval head under a reranking setting with 5 contriever candidates and 45 random negatives. I use the heads to get top-5 passages then generate the response. The setting may not be suitable here.}\mingda{does it tell you average heads is better? I think we probably need some results to support our choices here. It does not need to be perfect since these choices probably are made during preliminary experiments. If you don't have them, I can launch some quick runs for comparison. Same for the numbers of pseudo-positive/negative passages used in the retrieval warmup. Do you have any results or rationale that help you decide the exact values?}
Similarity between query and passage embeddings is computed via dot product:
\begin{align}
    s(q, p) = \mathbf{E}_q \cdot \mathbf{E}_p \label{eq:dot_score}
\end{align}

We choose to pool over query and key attention states based on the intuition that their dot product underlies the attention mechanism and is pretrained to capture token-wise relevance. By aggregating these signals across tokens, we aim to capture query-passage-level semantic relevance.

\paragraph{Middle Layers as Reader.}
The middle layer group $\mathcal{L_M}$ functions as a \textit{reader} by enabling cross-attention from the input query tokens to the retrieved passage tokens, thereby incorporating external information into the query representation. Given $k$ retrieved passages, we jointly encode the concatenation of all $k$ passages to form the key and value states for layers $b$ through $t$. \wenzheng{Maybe we can include the results of passage parallel encoding and mention it is cleaner but performs worse in Appendix?}\mingda{Sounds good!}
Cross-attention is then performed from the query’s attention states to these key and value states, allowing the model to read and integrate relevant content from the passages. This aligns with prior findings that middle layers of language models are particularly effective at attending to and integrating long-range contextual information~\citep{fang2024unimem,yang2024memory3}.

\paragraph{Top Layers Disable Cross-Attention.}
In the top layer group $\mathcal{L_T}$, we optionally disable cross-attention from the input query tokens to the retrieved passage tokens solely to reduce computational and memory overhead. This design choice is made for efficiency purposes; empirically, we find it results in only a minor performance drop when the layer boundary $t$ is properly tuned as a hyperparameter.

\paragraph{Position IDs.}
Language models using RoPE~\citep{su2024roformer} are highly sensitive to position IDs. To prevent interference between the query and passage position encodings during reading, we shift the query's position IDs to the right rather than starting from zero. Let $l_{max}$ denote the maximum passage length and $k$ the number of retrieved passages. We shift the query position IDs by $k\cdot l_{max}$ tokens to account for the total length. 

\subsection{Training}\label{subsec:train}
We train \thiswork using a multi-task objective that jointly optimizes generation and retrieval:

\begin{align}
J = J_{\text{gen}}(r \mid q, \mathcal{C}) + \lambda \cdot J_{\text{ret}}(q, \mathcal{C}) \label{eq:joint}
\end{align} 

Here, $J_{\text{gen}}(r \mid q, \mathcal{C})$ denotes the generation loss, implemented as the standard causal language modeling loss over the response tokens $r$, conditioned on the input query $q$ and a set of sampled candidate passages $\mathcal{C}$. The term $J_{\text{ret}}(q, \mathcal{C})$ denotes the retrieval loss, computed over the query $q$ and the same set of candidate passages $\mathcal{C}$, and is further detailed in the two-stage formulation described in Section~\ref{sec:ret}. The hyperparameter $\lambda$ balances the relative importance of the retrieval loss, allowing us to control the trade-off between retrieval accuracy and generation quality during training.

\subsubsection{Retrieval Objective}\label{sec:ret}

While the overall training objective remains consistent across both stages—combining generation and retrieval losses as in \eqref{eq:joint}—the retrieval loss component $J_{\mathrm{ret}}$ varies depending on the training phase. In this section, we describe the two-stage training process used to endow \thiswork with strong retrieval capabilities.

\paragraph{Warmup.}
Since the pretrained language model is not inherently optimized for retrieval, we begin with a warmup stage that introduces basic retrieval ability. We adopt a Multi-Label NCE loss~\citep{zhang2022entqa} as the retrieval objective and construct supervision using pseudo-labeled data generated by a strong off-the-shelf retriever, Contriever-MSMARCO~\citep{izacard2022unsupervised}. For each query $q$, we retrieve the top-5 passages as pseudo-positive examples, denoted by $\mathcal{P}(q)$. We then sample a small set of pseudo hard negatives, denoted by $\mathcal{N}_h(q)$ (e.g., $|\mathcal{N}_h(q)| < 10$), from passages ranked 10–50.\footnote{We find this approach effective in preliminary experiments, though we did not perform extensive hyperparameter tuning.} While these passages may still be somewhat relevant, they are less likely to contain the key information necessary to answer the query. This selection introduces meaningful retrieval difficulty. \karl{Some justification of why top 5 are positive but top 10-50 are negative would be good, to mitigate the worry that these are spurious negatives.} \wenzheng{Thanks for pointing out this! I included some justification for this. Does it look good?} We also use in-batch negatives across devices as additional random negatives $\mathcal{N}_r(q)$. The full negative set is $\mathcal{N}(q) = \mathcal{N}_h(q) \cup \mathcal{N}_r(q)$, and the candidate set is $\mathcal{C} = \mathcal{P}(q) \cup \mathcal{N}(q)$. The retrieval loss for this stage is defined as:

\begin{align}
\scalebox{0.81}{$
J_{\mathrm{ret}}(q,\mathcal{C}) = -\sum_{p \in \mathcal{P}(q)}
\log \left( \frac{\exp(s(q,p))}{\exp(s(q,p)) + \sum_{p' \in \mathcal{N}(q)} \exp(s(q, p'))} \right)
$}
\label{eq:mNCE}
\end{align}

\paragraph{Self-Distillation.}
To further enhance retrieval performance, we employ language model perplexity distillation~\citep{izacard2023atlas}, which assesses how much each candidate passage improves the language model's likelihood of generating the ground-truth response, conditioned on the query. Specifically, for each candidate passage $p \in \mathcal{C}$, we compute the log-likelihood of the gold response $r$ given the concatenation of $p$ and $q$, denoted as $\log P_{\text{LM}}(r \mid p, q)$. This defines a soft target distribution over candidate passages:\karl{Isn't this just renormalizing $P_{\text{LM}}(r|p,q)$? I.e., exp and log cancel.} \wenzheng{Good catch! I just followed Atlas paper to write this equation and I think this looks clearer than the renormalization objective.}

\begin{align}
P_{T}(p \mid q, r) = \frac{\exp(\log P_{\text{LM}}(r \mid p, q)/\tau_t)}{\sum_{p' \in \mathcal{C}} \exp(\log P_{\text{LM}}(r \mid p', q)/\tau_t)}
\label{eq:ppl_prob}
\end{align}

 We also define the retrieval model’s predicted distribution based on the similarity scores:

\begin{align}
P_{R}(p \mid q) = \frac{\exp(s(q, p)/\tau_r)}{\sum_{p' \in \mathcal{C}} \exp(s(q, p')/\tau_r)}
\label{eq:ret_prob}
\end{align}

The retrieval loss is then computed as the KL divergence between the target and predicted distributions:

\begin{align}
J_{\mathrm{ret}}(q, \mathcal{C}) = \mathrm{KL} \left( \overline{P_{T}(p \mid q, r)} \parallel P_{R}(p \mid q) \right)
\label{eq:ppl_loss}
\end{align}

Here, $\overline{P_{T}(p \mid q, r)}$ indicates that gradients are not backpropagated through the target distribution. Note that this stage also involves joint training; the only difference from the warmup phase lies in the retrieval loss $J_{\mathrm{ret}}$.
\karl{Is the distillation happening after the joint training is done, or during? Might be good to clarify that.} \wenzheng{Thanks! I included the last sentence to clarify this.}

\subsection{Inference}\label{subsec:infer}
At inference time, we first embed all passages in the knowledge corpus using the bottom layer group $\mathcal{L}_B$ of the model. These embeddings are stored in an approximate nearest neighbor (ANN) index (e.g., FAISS~\citep{douze2024faiss}) hosted on a remote server for efficient retrieval. 

As illustrated in Figure~\ref{fig:arch}, given a query, the \thiswork model performs the following steps to generate a response:

\begin{enumerate}
\item The bottom layers $\mathcal{L}_B$ encode the input query and generate a query embedding, which is sent to the remote ANN search server.
\item The ANN server retrieves the top-$k$ most relevant passages based on the query embedding and returns their passage IDs.
\item The middle layers $\mathcal{L}_M$ continue processing the information by applying cross-attention to the KV states of the retrieved passages.
\item The top layers $\mathcal{L}_T$ complete the encoding and decoding process without cross-attention, generating the next token.
\item The above steps are repeated at each decoding step. Notably, the query embeddings are computed only once at the end of the input prompt, and passage retrieval is not re-triggered thereafter.\footnote{While \thiswork is general and can be adapted for iterative retrieval, we intend to focus this work on the single retrieval setup and will leave iterative retrieval for future work.} In subsequent decoding steps, cross-attention continues to use the cached key-value states, and this process repeats until the model reaches a stopping criterion (e.g., an end-of-sequence token).\wenzheng{I mentioned the retrieval occurs only once here. Does it sound good?}\mingda{Thank you! I've added a footnote for clarification. Quick question: Will the retrieved passages be visible to the bottom layers in the future decoding steps?}\wenzheng{No, only visible to the middle layers in the future decoding steps.}\karl{Also this is consistent with training, since you use the query to get candidates?}
\end{enumerate}

\section{Experiment}
\subsection{Experimental Setup}

\paragraph{Training.} For training, we consider two types of datasets: (1) datasets requiring retrieval knowledge: \nq (\nqshort;~\citealp{kwiatkowski-etal-2019-natural}) and \hotpot (\hotpotshort;~\citealp{yang-etal-2018-hotpotqa}); and (2) datasets without requiring retrieval knowledge, where we use the instruction tuning datasets from \citet{lin2024radit} (see Appendix~\ref{app:instruct_tuning_datasets} for a complete list of these datasets). Inspired by \citet{chen-etal-2022-improving}, we also incorporate two synthetic, retrieval-free tasks into the training to enhance instruction-following capabilities: phrase denoising, and next/previous sentence generation. The training data for phrase denoising is generated by prompting LLMs (we use Llama-3.1 70B) with a paragraph from Wikipedia. For the sentence generation task, we construct it randomly using content from Wikipedia.

For all these datasets, we use a subset of 5,000 examples from their training splits in each dataset. In addition, we use 1,000 examples from the \nqshort dev split as the development set. We use the December 2021 Wikipedia from \citet{izacard2023atlas} as our knowledge corpus. Additionally, we spend approximately 10\% of training on plain text from Wikipedia to prevent models from overfitting to the downstream tasks.

\begin{table*}[ht]
    \centering\small
    \begin{adjustbox}{width=\textwidth}
    \begin{tabular}{p{0.40\textwidth}p{0.60\textwidth}}
    \hline
    Task &  Template  \\
    \hline
    \multicolumn{2}{l}{\emph{Knowledge-Intensive Tasks}} \\
    \nqshort, \hotpotshort, \simpleqashort, \wikiqashort & {Q:} \{question\} {A:} \{answer\} \\
    \aida & \{context\} {Output the Wikipedia page title of the entity mentioned between [START] and [END] in the given text} {A:} \{answer\} \\
    \fevershort & {Is this statement true?} \{statement\} {A:} \{answer\} \\
    \trex, \zsre & \{entity\} [SEP] \{relation\} Provide the answer corresponding to the relation specified after [SEP] for the entity mentioned before [SEP] {A:} \{answer\}\\
    \hline
    \multicolumn{2}{l}{\emph{Instruction-Tuning Tasks}} \\
    Dialogue Completion & \{turn$_1$\} \{turn$_2$\} \{turn$_3$\} ... \\
    Reading Comprehension & \{context\} Q: \{question\} A: \{answer\} \\
    Summarization &  \{context\} Summarize this article: \{summary\} \\
    Phrase Denoising & \{context\} Recover the original phrases marked between [START] and [END] in the given text A: \{answer\} \\
    Sentence Generation & \{context\} [SEP] next/previous sentence Generate a sentence corresponding to the relation specified after [SEP] for the context mentioned before [SEP] A: \{sentence\} \\\hline
    \end{tabular}
    \end{adjustbox}
    \caption{Prompt templates. We only use retrieval for knowledge-intensive tasks. For simplicity, we list task categories for a subset of the instruction tuning datasets. See Appendix~\ref{app:instruct_tuning_datasets} for more detailed description.}
    \label{tab:task_template}
\end{table*}

\begin{table*}[t]
    \centering\small\setlength{\tabcolsep}{0.3em}
    \begin{adjustbox}{width=\textwidth}
    \begin{tabular}{l|c|c|c|c|c|c|c|c|c}
                 & \nqshort & \simpleqashort & \hotpotshort & \wikiqashort & \trex & \zsre & \fevershort & \aida & avg  \\\hline
    Llama-3.2$_{\text{3B}}$ \\
       +\rait & 43.2 (77.0) & 38.1 (48.2) & 35.9 (48.8) & 33.4 (43.3) & 54.2 (84.3) & 58.1 (86.6) & 79.2 (-) & 40.1 (38.1) & 47.8 (60.9) \\
       +\radit & 43.4 (77.5) & 38.8 (48.7) & 36.4 (49.3) & 34.0 (43.5) & 55.0 (85.0) & 59.0 (87.2) & 80.5 (-) & 41.0 (38.2) & 48.5 (61.3) \\
       +\raditllama & 43.9 (78.0) & 39.8 \bf{(49.9)} & 37.0 (49.8) & 35.1 (44.0) & 55.8 (85.9) & 60.0 (87.9) & 80.2 (-) & 41.1 (38.4) & 50.4 (64.0) \\
       +\thiswork & \bf 44.1 (78.4) & \bf 40.3 (50.0) & \bf 37.3 (50.2) & \bf 35.5 (44.5) & \bf 60.8 (90.2) & \bf 65.4 (93.2) & \bf 83.8 (-) & \bf 52.6 (58.3) & \bf 52.5 (66.4) \\\hline
    Llama-3.1$_{\text{8B}}$ \\
       +\rait & 45.1 (77.0) & 39.0 (48.2) & 36.9 (48.8) & 34.4 (43.3) & 55.0 (84.3) & 59.1 (86.6) & 83.2 (-) & 41.1 (38.1) & 49.2 (60.9)\\
       +\radit & 45.7 (77.7) & 38.9 (48.9) & 37.2 (49.1) & 34.9 (44.0) & 56.1 (85.4) & 60.1 (87.8) & 85.1 (-) & 41.5 (38.8) & 49.9 (61.7) \\
       +\raditllama & 46.1 (78.7) & 40.7 (50.3) & 37.9 (50.2) & 35.6 (44.8) & 57.0 (86.1) & 61.2 (88.1) & 86.2 (-) & 42.1 (39.2) & 50.9 (62.5) \\
       +\thiswork & \bf 46.4 (79.1) & \bf 41.3 (51.2) & \bf 38.4 (50.9) & \bf 36.0 (45.2) & \bf 62.5 (92.7) & \bf 67.1 (94.0) & \bf 89.2 (-) & \bf 54.2 (62.4) & \bf 54.4 (67.9) \\\hline
    \end{tabular}
    \end{adjustbox}
    \caption{Evaluation results for 8 knowledge-intensive tasks. We report exact match scores for generation tasks and retrieval recall (shown in parentheses) for retrieval tasks. Retrieval recall is not reported for \fevershort, as it is a classification task. All these methods use retrieval augmentation.
    }
    \label{tab:main_result}
\end{table*}

\paragraph{Evaluation.} We evaluate models on 8 different knowledge-intensive tasks to assess their various capabilities, specifically: 
\begin{itemizesquish}
    \item Basic question answering: \nqshort, \simpleqa~(\simpleqashort;~\citealp{wei2024measuring});
    \item Multihop reasoning: \hotpotshort, \wikiqa~(\wikiqashort;~\citealp{ho-etal-2020-constructing});
    \item Instruction following: (1) relation extraction: \trex~\citep{elsahar-etal-2018-rex}, \zsre~\citep{levy-etal-2017-zero}, (2) fact checking: \fever~(\fevershort;~\citealp{thorne-etal-2018-fever}), and (3) entity linking: \aida~\citep{hoffart-etal-2011-robust}.
\end{itemizesquish}

For all these datasets, we report exact matches as the evaluation metric for generation tasks and recall rates for retrieval tasks. The retrieval recall is measured by the percentage of instances where the top-retrieved results contain the answers as substrings.
We omit retrieval recall for \fevershort as it is a classification task where the answer strings are either ``True'' or ``False''.

For \hotpotshort, \trex, \zsre, \fevershort, and \aida, we use development sets from the KILT benchmark~\citep{NEURIPS2020_6b493230}. For \simpleqashort and \nqshort, we use the official test set. For \wikiqashort, we use their development set. For all datasets, we utilize the entire input prompts as queries for the retrievers. 
We describe our task templates in Table~\ref{tab:task_template}.

\paragraph{Baselines.} We consider 3 baseline models:

\begin{itemizesquish}
    \item Retrieval Augmented Instruction Tuning (\rait): This approach involves directly incorporating retrieved passages from Contriever-MSMARCO
    into the context and fine-tuning the language models (LMs) on the training data;
    \item Retrieval Augmented Dual Instruction Tuning (\radit; \citealp{lin2024radit}): In this method, we first fine-tune the Contriever-MSMARCO on the training subsets of \nqshort and \hotpot using Equation \ref{eq:ppl_loss}. Subsequently, we perform fine-tuning as in \rait, utilizing the fine-tuned retriever;
    \item \radit with Llama as the Retriever (\raditllama): Here, we replace the Contriever used in \radit with the first 8 layers from the Llama models.\footnote{We choose to use first 8 layers for fair comparison as \thiswork uses the same layers for retrieval.} To ensure effective retrieval performance, we initially warm up the Llama retrievers with pseudo labels generated by Contriever-MSMARCO using Equation \ref{eq:mNCE}.
\end{itemizesquish}

\paragraph{Hyperparameters.} We use Llama-3.2 3B and Llama-3.1 8B as the base models for \thiswork. For both models, the layer boundary $b$ is set to 7.\footnote{Since we label the first layer of a LLM as layer 0, a layer boundary $b$ of 7 means that the bottom layer group contains the first 8 layers.} For Llama-3.2 3B, the layer boundary $t$ is 19, while for Llama-3.1 8B, it is 23. We train for 10 epochs and perform the retrieval warmup in the first 3 epochs. When retrieving passages, we take the top 10 most relevant documents.

See Appendix~\ref{app:param_and_compute} for more details on the baselines and computational resources.

\begin{figure*}[t]
    \centering
    \includegraphics[width=0.5\linewidth]{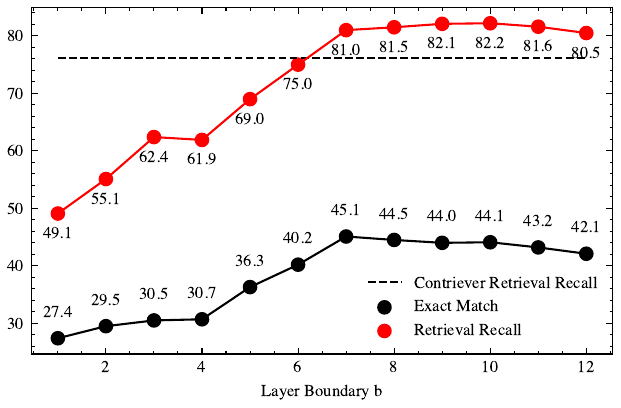}%
    \includegraphics[width=0.5\linewidth]{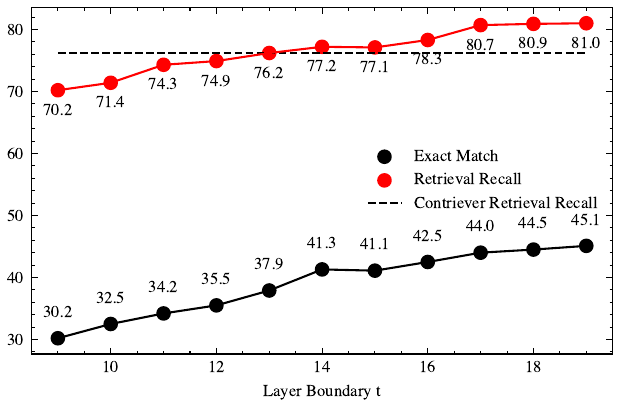}
    \caption{Exact match and retrieval recall on the NQ dev set using Llama-3.2 3B with different values of $b$ (left side) and $t$ (right side). When varying one layer boundary, we keep the other constant.}
    \label{fig:layer_boundary_ablation}
\end{figure*}

\begin{table*}[t]
    \centering\small\setlength{\tabcolsep}{0.3em}
    \begin{adjustbox}{width=\textwidth}
    \begin{tabular}{l|c|c|c|c|c|c|c|c|c}
                 & \nqshort & \simpleqashort & \hotpotshort & \wikiqashort & \trex & \zsre & \fevershort & \aida & avg  \\\hline
        self-distillation only & 29.9 (61.2) & 30.1 (39.9) & 29.8 (41.9) & 27.5 (37.4) & 35.6 (64.9) & 40.9 (65.9) & 67.7 (-) & 28.3 (22.5) & 36.2 (47.7) \\
        warmup only & \bf 44.0 (78.3) & 39.9 \bf{(50.0)} & 37.1 (50.0) & 35.1 (44.2) & 56.5 (87.0) & 61.2 (88.3) & 81.0 (-) & 45.2 (42.9) & 50.0 (63.0) \\
        warmup+self-distillation & \bf 44.1 (78.4) & \bf 40.3 (50.0) & \bf 37.3 (50.2) & \bf 35.5 (44.5) & \bf 60.8 (90.2) & \bf 65.4 (93.2) & \bf 83.8 (-) & \bf 52.6 (58.3) & \bf 52.5 (66.4) \\\hline
    \end{tabular}
    \end{adjustbox}
\caption{Exact match scores and retrieval recall (shown in parentheses) for \thiswork using Llama-3.2 3B as the base model, trained with different retrieval objectives.}\label{tab:retrieval_objective_ablation}
\end{table*}

\subsection{Experimental Result}

\begin{table*}[t]
    \centering
    \begin{tabular}{l|c|c|c|c|c}
         & \trex & \zsre & \fevershort & \aida & avg\\\hline
      No templates & 55.8 (85.9) & 60.0 (87.9) & 80.2 (-) & 41.1 (38.4) & 59.3 (70.7) \\
      Oracle templates &\bf 61.4 (90.7) & \bf 66.0 (93.6) & \bf 83.9 (-) & \bf 66.1 (72.3) & \bf 69.4 (85.5)  \\
      \thiswork & 60.8 (90.2) & \bf 65.9 (93.5) & \bf 83.8 (-) & 52.6 (58.3) & 65.8 (80.7) \\\hline
    \end{tabular}
    \caption{Exact match scores and retrieval recall (shown in parentheses) for \raditllama using Llama-3.2 3B as the base model, evaluated with various query templates. In the case of ``no templates'', the inputs to the LLMs are used directly as queries.}
    \label{tab:query_template_ablation}
\end{table*}

\begin{table*}[t]
\centering\small\setlength{\tabcolsep}{0.3em}
    \begin{tabular}{l|c|c|c|c|c|c|c|c|c}
                 & \nqshort & \simpleqashort & \hotpotshort & \wikiqashort & \trex & \zsre & \fevershort & \aida & avg  \\\hline
       \thiswork  & \bf 44.1 (78.4) & \bf 40.3 (50.0) & \bf 37.3 (50.2) & \bf 35.5 (44.5) & \bf 60.8 (90.2) & \bf 65.4 (93.2) & \bf 83.8 (-) & \bf 52.6 (58.3) & \bf 52.5 (66.4) \\
      w/o all IT tasks & 42.9 (76.4) & 38.1 (48.2) & 35.2 (47.7) & 33.7 (42.0) & 43.5 (69.3) & 49.5 (70.2) & 76.2 (-) & 25.4 (20.5) & 43.1 (53.5) \\
      w/o PD and SG & \bf 44.0 (78.5) & \bf 40.1 (50.1) & \bf 37.4 (50.3) & \bf 35.4 (44.4) & 53.3 (82.8) & 57.1 (84.9) & 81.2 (-) & 40.5 (41.2) & 48.6 (61.7) \\
    \end{tabular}
    \caption{Exact match scores and retrieval recall (shown in parentheses) for \thiswork using Llama-3.2 3B as the base model, trained with different combinations of instruction tuning datasets. IT tasks refer to instruction tuning tasks, PD stands for phrase denoising, and SG denotes sentence generation.}\label{tab:instruct_tuning_ablation}
\end{table*}

Table \ref{tab:main_result} presents our main evaluation results.
Each model variant—\rait, \radit, \raditllama, and \thiswork—exhibits different performance levels, with \thiswork consistently achieving the highest scores across all tasks, including both in-domain and unseen tasks. Notably, the improvements on unseen tasks are more substantial, suggesting that \thiswork effectively enhances cross-task generalization. For Llama-3.2 3B, the average exact match score increases from 47.8 with \rait to 52.5 with \thiswork, while for Llama-3.1 8B, the score rises from 49.2 to 54.4. Although \radit shows improvements over \rait, \thiswork further enhances performance. Crucially, \thiswork significantly outperforms \raditllama, indicating that the improvements are not merely due to using a more powerful base model (i.e., the first 8 layers of Llama models) for retrieval. Importantly, the enhancements are evident in both exact match scores and retrieval recalls, demonstrating that \thiswork improves both generation quality and retrieval performance. It is worth noting that compared to the baseline approaches, the most substantial improvements with \thiswork are seen in tasks that queries need to be formulated more differently from input prompts, such as \trex, \zsre, \fever, and \aida. Among these tasks, \aida shows the most significant improvements, with over a 20-point increase in retrieval recall and more than a 10-point rise in exact match scores for both Llama-3.1 3B and Llama-3.1 8B, likely due to the inadequacy of directly using input prompts as queries in \aida. This underscores \thiswork's effectiveness in formulating implicit queries and embedding instruction-following capabilities into retrievers. Overall, these results demonstrate that \thiswork significantly enhances the models' ability to accurately retrieve and apply knowledge, with improvements more significant in tasks requiring diverse formats.
\section{Analysis}

\subsection{Layer Group Boundary Ablation}

In this section, we examine the effects of the layer boundaries $b$ and $t$. The findings are presented in Figure~\ref{fig:layer_boundary_ablation}. To facilitate comparison, we vary one layer boundary while keeping the other constant. We note that increasing $b$ reduces the number of layers allocated to the middle layer group, which includes layers for reading and generation. Conversely, increasing $t$ does not affect the retrieval layers. Overall, we find that increasing $b$ enhances retrieval recall, with improvements leveling off once $b$ reaches 7. This plateau is likely due to diminished generation performance, which results in less precise training signals for self-distillation. This underscores the importance of balancing parameters between retrieval and generation. On the other hand, as expected, increasing $t$ consistently yields improvements. Although these improvements seem to plateau at 19, we refrain from further increasing $t$ primarily due to memory constraints. We plan to leave more memory-efficient training of \thiswork for future exploration.

\subsection{Retrieval Objective Ablation}

We conduct experiments to compare the effects of different retrieval training objectives. The results are presented in Table~\ref{tab:retrieval_objective_ablation}. During training, we consistently apply each retrieval objective throughout the entire process. For instance, in the "warmup only" experiment, we extend the use of the warmup objective to 10 epochs instead of limiting it to the initial 3 epochs. Our findings indicate that the warmup objective provides a baseline performance across all tasks and is particularly beneficial for tasks with direct supervision. Self-distillation builds on this baseline, further enhancing model performance on unseen test tasks. Overall, the two training objectives complement each other effectively.

\subsection{Effect of Query Templates}

We also examine the impact of using different query templates for the baseline approach, \raditllama. The results are detailed in Table~\ref{tab:query_template_ablation}. In these experiments, we omit QA tasks because their ``no templates'' and ``oracle templates'' setups are almost the same. Overall, ``oracle templates'' still provides the best performance. The improvements are particularly notable on \aida. However, it is important to highlight that \thiswork achieves highly competitive performance on 3 out of 4 tasks and already shows significant improvement on the remaining task compared to using ``no templates.''




\subsection{Effect of Instruction Tuning for Retrieval}

In Table~\ref{tab:instruct_tuning_ablation}, we explore the effects of training on instruction tuning datasets. The table shows that omitting all instruction tuning datasets leads to a decline in model performance on both in-domain tasks (\nqshort, \simpleqashort, \hotpotshort, and \wikiqashort) and out-of-domain tasks. Notably, removing only phrase denoising and sentence generation has a minimal impact on in-domain tasks but causes more pronounced negative effects on out-of-domain tasks, except for \fevershort. This exception likely arises because \fevershort's task format is more similar to the in-domain tasks than other tasks. This suggests that instruction tuning tasks aid models in understanding task formats, and \thiswork can transfer this knowledge from generation to retrieval due to its unified model architecture.

\section{Conclusion}
We present ImpRAG, a query-free retrieval-augmented generation (RAG) system that implicitly captures information needs without requiring human-specified queries. Unlike prior work that treats retrieval and generation as separate components with independently trained models, ImpRAG unifies them within a single decoder-only language model by partitioning it into specialized layer groups and jointly optimizing for both retrieval and generation.
The same model parameters and forward pass are shared across retrieval and generation, effectively minimizing the mismatch between the retriever and the generator. ImpRAG demonstrates strong performance across eight knowledge-intensive tasks, outperforming traditional RAG systems and delivering substantial gains on unseen tasks with diverse formats.
\section{Limitations}
One limitation of this work is its focus on a single-pass retrieval setup; we do not explore iterative or multi-hop retrieval, which could further enhance performance on complex reasoning tasks. Adapting ImpRAG to iterative retrieval remains an important direction for future work.

Another limitation of our method is the increased structural complexity of the model. However, it does not add to implementation or fine-tuning complexity compared to prior joint RAG approaches like \radit, which typically require iterative updates to both the retriever and generator. Moreover, our inference process remains simple and efficient: it avoids task-specific retrieval templates and separate retriever–generator pipelines by unifying retrieval and generation into a single forward pass, reducing engineering overhead and simplifying deployment.

Our method is also evaluated exclusively using the LLaMA 3 family of models. While the approach is broadly applicable, its generalizability to other architectures and model sizes has yet to be validated.

Additionally, the warmup training stage relies on pseudo-labeled data generated by Contriever-MSMARCO. Although this provides a strong starting point, we expect that using more powerful retrievers or human-labeled data could lead to further gains by offering higher-quality supervision early in training. To ensure broad applicability to datasets without gold annotations, we deliberately use trained retrievers rather than supervised signals. Notably, our evaluation does not require any additional trained retrievers from external sources.


\bibliography{anthology,custom}

\appendix

\section{Passage Encoding}\label{app:psg_enc}

Given $k$ retrieved passages, we must obtain the key-value (KV) states from the middle layer group $\mathcal{L}_M$ to enable cross-attention. We explore three passage encoding strategies, summarized in Table~\ref{tab:psg_enc}.

First, we consider Independent Encoding, where each passage is encoded separately using position IDs starting from zero, following the parallel encoding strategy in~\citet{yen2024long}. The resulting KV states are then concatenated across passages.

Second, we examine Concatenated Encoding (Segmented), in which passages are concatenated into a single sequence, but attention across passages is blocked to prevent inter-passage interaction.

Third, we evaluate Concatenated Encoding (Full Attention), where passages are concatenated and full cross-passage attention is allowed throughout the encoding.

We conduct these experiments by finetuning Llama-3.1 8B model on the Natural Questions (NQ) dataset using the top-10 passages retrieved by Contriever-MSMARCO, and report Exact Match (EM) scores on the development set. As shown in Table~\ref{tab:psg_enc}, the two simpler strategies—Independent Encoding and Segmented Concatenation—perform similarly, while Full Attention Concatenation yields a clear performance improvement, highlighting the benefit of modeling inter-passage dependencies.

\begin{table}[ht]
\centering
\begin{tabular}{c|c}
\hline
Encoding Method & Dev EM \\
\hline\\[-1em]
Independent Encoding & 51.7 \\
Segmented Concatenation & 51.4 \\
Full Attention Concatenation & 53.3 \\
\hline
\end{tabular}
\caption{Performance of different passage encoding strategies.}
\label{tab:psg_enc}
\end{table}

\section{Freezing Passage Representations}\label{app:freeze}

We investigate the impact of freezing passage representations—either hidden states or key-value (KV) states—during inference with a fixed retriever. All experiments are conducted using a fine-tuned LLaMA-3.1 8B model and the top-10 passages retrieved by Contriever-MSMARCO on the Natural Questions (NQ) dataset. Results are reported in Table~\ref{tab:freeze}. 

We explore two freezing strategies, both using the Independent Encoding approach described in Appendix~\ref{app:psg_enc}. In the first variant, Frozen Hidden States, we freeze the hidden representations of retrieved passages as produced by the initial (untrained) LLaMA-3.1 8B model, and pass them through the trained key/value projection layers to generate the KV states used in cross-attention.

In the second variant, Frozen KV States, we directly freeze the key and value attention states of the passages, also obtained from the initial LLama-3.1 8B model.

We observe that both freezing methods yield comparable performance, slightly underperforming the fully dynamic setting where passage KV states are computed using the trained model.

\begin{table}[ht]
\centering
\begin{tabular}{c|c}
\hline
Method & Dev EM \\
\hline\\[-1em]
No Freezing & 51.4 \\
Frozen Hidden States & 50.8 \\
Frozen KV States & 50.7 \\
\hline
\end{tabular}
\caption{Performance of freezing different passage representations on NQ dev set with top-10 Contriever-MSMARCO retrieved passages.}
\label{tab:freeze}
\end{table}

\section{Passage KV States Compression}\label{app:compress}
When we use independent encoding strategy in Appendix~\ref{app:psg_enc}, one benefit will be that we can save the middle layer group key value states for all the passages in knowledge corpus in disk and during inference after retrieval we can load the key value states from disk without recomputation. However, this will result in a large amount of disk spaces. Thus, we consider two compression strategies: token compression and product quantization and we conduct experiments following the same setting as the Frozen KV states in Appendix~\ref{app:freeze}. Specifically, take for token compression, we use the Heavy Hitter~\citep{zhang2023h2o} and only keep half number of tokens for each passage. For production quantization, we use FAISS codec with index type OPQ32x128-PQ32x8 for each key value head, which is trained on 500k randomly sampled wikipedia passages. The compression rate with this quantization is $\frac{128\times 2}{32}=8$ for original bfloat16 state vector of each attention head. We report the results in Table~\ref{tab:compress}. We can see that both strategies don't hurt the performance much.

\begin{table}[ht]
    \centering
    \begin{tabular}{c|c}
    \hline 
     Compression   &  Dev EM \\
     \hline\\[-1em]
      No Compression   &  50.7 \\
      Heavy Hitter & 49.9 \\
      Product Quantization & 50.3 \\
      \hline
    \end{tabular}
    \caption{The results for various compression techniques.}
    \label{tab:compress}
\end{table}

\section{Instruction-Tuning Datasets}\label{app:instruct_tuning_datasets}
We use OpenAssistant Conversations Dataset~(\oasst;~\citealp{kpf2023openassistant}), Conversational Question Answering~(\coqa;~\citealp{reddy-etal-2019-coqa}), Discrete Reasoning Over Paragraphs~(\drop;~\citealp{dua-etal-2019-drop}), \newsqa~\citep{trischler-etal-2017-newsqa}, \pubmedqa~\citep{jin-etal-2019-pubmedqa}, QA for Artificial Intelligence~(\quail;~\citealp{Rogers_Kovaleva_Downey_Rumshisky_2020}), SQuAD v2~\citep{rajpurkar-etal-2018-know},\footnote{We only use answerable questions from SQuAD v2.} and CNN DailyMail~\citep{chen-etal-2016-thorough} The templates for these datasets are shown in Table~\ref{app_tab:task_template}.

\begin{table}[ht]
    \centering
    \begin{adjustbox}{width=\columnwidth}
    \begin{tabular}{p{0.40\textwidth}p{0.60\textwidth}}
    \hline
    Task &  Template  \\
    \hline
    \multicolumn{2}{l}{\emph{Instruction-Tuning Tasks}} \\
    \oasst & \{turn$_1$\} \{turn$_2$\} \{turn$_3$\} ... \\
    \coqa, \drop, \newsqa, \pubmedqa, \squad & \{context\} Q: \{question\} A: \{answer\} \\
    \cnndaily &  \{context\} Summarize this article: \{summary\} \\
    \end{tabular}
    \end{adjustbox}
    \caption{Prompt templates. We only use retrieval for knowledge-intensive tasks.}
    \label{app_tab:task_template}
\end{table}

\section{Baselines and Computational Resources}\label{app:param_and_compute}

\paragraph{Discussions on Baselines.} For all these baselines, we use the retrieve-then-generate paradigm, i.e., begin by retrieving candidates using the retrievers and then incorporate them into the context for training and inference. This implies that these baselines require an additional retriever, leading to increased computational costs and a higher number of model parameters compared to \thiswork. However, since this is a standard practice for retrieval-augmented models, we continue to use them in the baselines to establish stronger comparisons. 

\paragraph{Computational Resources.} We use NVIDIA H100 GPUs. Each training session requires 8 H100 GPUs, and hosting the index also demands an additional 8 GPUs. Training the baseline approaches takes roughly 96 GPU hours, whereas our models require approximately 160 GPU hours.

\end{document}